\documentclass{article}

    \PassOptionsToPackage{numbers, compress}{natbib}

    \usepackage[preprint]{neurips_2024}

\usepackage[utf8]{inputenc} 
\usepackage[T1]{fontenc}    
\usepackage{hyperref}       
\hypersetup{colorlinks=false, allcolors=black, pdfborder={0 0 0}}
\usepackage{url}            
\usepackage{booktabs}       
\usepackage{amsfonts}       
\usepackage{nicefrac}       
\usepackage{microtype}      
\usepackage{xcolor}         

\usepackage{graphicx}
\usepackage{bm}
\usepackage{amsmath}
\usepackage[capitalize,noabbrev]{cleveref}
\usepackage{subcaption}
\usepackage{multirow}
\usepackage{siunitx}
\usepackage{wrapfig}
\usepackage{makecell}
\usepackage{sidecap}

\newcommand*{\img}[1]{%
    \raisebox{-.3\baselineskip}{%
        \includegraphics[
        height=\baselineskip,
        width=\baselineskip,
        keepaspectratio,
        ]{#1}%
    }%
}
\usepackage{algorithm}
\usepackage{algorithmic}

\title{Prospective Messaging: \\Learning in Networks with Communication Delays}

\author{%
  Ryan Fayyazi$^*$, Christian Weilbach$^*$, Frank Wood\\
  Department of Computer Science\\
  University of British Columbia\\
  Vancouver, Canada\\
  \texttt{\{rfayyazi,weilbach,fwood\}@cs.ubc.ca} \\
}

\begin{document}

\newcommand{\R}{\mathbb{R}}

\newcommand{\delay}[2]{\delta_{#2#1}}
\newcommand{\ddelay}[1]{\bm{\delta}_{#1}}

\newcommand{\ubreve}{\breve{\mathbf{u}}}

\newcommand{\usc}[2]{u_{#1}(#2)} 
\newcommand{\ud}[3]{\Bar{u}_{#2#1}(#3)}
\newcommand{\up}[2]{\breve{u}_{#1}(#2)}
\newcommand{\usout}[1]{\mathbf{u}_{out}(#1)}

\newcommand{\vs}[2]{e_{#1}(#2)} 
\newcommand{\vd}[3]{\Bar{e}_{#2#1}(#3)}

\newcommand{\zs}[2]{\mathbf{s}_{#1}(#2)}
\newcommand{\zd}[2]{\Bar{\mathbf{s}}_{#1}(#2)}

\newcommand{\usv}[2]{\dot{u}_{#1}(#2)}
\newcommand{\upv}[2]{\dot{\breve{u}}_{#1}(#2)}

\newcommand{\usa}[2]{\ddot{u}_{#1}(#2)}

\newcommand{\Wc}[3]{W_{#2#1}(#3)}
\newcommand{\Wv}[3]{\dot{W}_{#2#1}(#3)}
\newcommand{\bi}[2]{b_{#1}(#2)}
\newcommand{\biv}[2]{\dot{b}_{#1}(#2)}

\newcommand{\ts}[1]{t_{#1}^{s}}
\newcommand{\te}[1]{t_{#1}^{e}}

\newcommand{\hist}[2]{\mathbf{h}_{#1}(#2)}
\newcommand{\netin}[2]{\mathbf{h}^{in}_{#1}(#2)}
\newcommand{\f}[1]{f_{#1}}

\newcommand{\params}{\mathbf{\theta}}   

\maketitle
\def\thefootnote{*}\footnotetext{These authors contributed equally to this work} 

\begin{abstract}
    Inter-neuron communication delays are ubiquitous in physically realized neural networks such as biological neural circuits and neuromorphic hardware. These delays have significant and often disruptive consequences on network dynamics during training and inference. It is therefore essential that communication delays be accounted for, both in computational models of biological neural networks and in large-scale neuromorphic systems. Nonetheless, communication delays have yet to be comprehensively addressed in either domain. In this paper, we first show that delays prevent state-of-the-art continuous-time neural networks called Latent Equilibrium (LE) networks from learning even simple tasks despite significant overparameterization. We then propose to compensate for communication delays by predicting future signals based on currently available ones. This conceptually straightforward approach, which we call prospective messaging (PM), uses only neuron-local information, and is flexible in terms of memory and computation requirements. We demonstrate that incorporating PM into delayed LE networks prevents reaction lags, and facilitates successful learning on Fourier synthesis and autoregressive video prediction tasks.
\end{abstract}

\section{Introduction}
\label{sec:intro}

Communication delays are inevitable in physically realized networks, since no signal can be transmitted faster than the speed of light. In biological neural networks (NNs) for example, inter-neuron communication delays range from less than one millisecond (ms) to over 100 ms \citep{aston1985, ferraina2002}.
As nervous system grow in size, these delays can place ecologically significant limitations on reaction speeds \citep{more2018}.
Recent work has also shown that delays as small as 20 ms in sensorimotor pathways can entirely disrupt models of relatively simple motor behaviour in small animals \citep{karashchuk2024}.
Neuromorphic hardware inherits the inter-neuron communication delays of biological NNs, from which they draw close inspiration \citep{schuman2022}. In this case delays are often programmable, and can be made much smaller than those found in biological NNs. However, since it is impossible to eliminate delays altogether, they pose a fundamental limitation on the size of these networks \citep{shainline2021}. Analogous communication delays also exist in digital computer chips, where they limit processing speeds. Although the theoretical limits on inter-transistors distances are far from being reached \citep{ossiander2022}, inter-chip communication delays are currently the major bottleneck for distributed memory access and processing on all computational scales \citep{mutlu2023, archer2019}. Finally communication delays play a crucial limiting role in multi-agent systems, such as a distributed group of robots and/or humans where messages need to be distributed in a peer-to-peer fashion \citep{tian2008, lin2016, yuan2023}.

This paper considers the especially challenging case where communication delays exist within a system undergoing learning. In particular, we focus on continuous-time NNs (CTNNs), i.e NNs wherein all neurons and parameters execute their own dynamics in parallel and continuously through time, according to a set of differential equations \citep{indiveri2011, hahne2017}. This setting is closely reflected in computational models of biological NNs, as well as neuromorphic hardware \citep{shrestha2022}. Nonetheless, the methods introduced in this paper are not specific to this setting and can in principle be applied to other computational processes with communication delays. Crucially, in contrast to efforts in computational neuroscience and distributed deep learning to accommodate for delays at specific points within circuits of interest, we address the most general setting in which all inter-neuron communications in the entire networks incur delays. 

Before continuing, we would like to offer some more intuition about the undesirable consequences of communication delays on NN learning and inference. Communication delays clearly limit an NN's reaction time, impeding performance in rapidly changing environments. If learning depends on the relationship between the network's output and an external instructive signal, response latency can disrupt learning as well. In particular, if the environment changes before information about the previous stimulus has time to propagate through the network, the network's output will be lagged with respect to any instructive signal in the new environmental state. 

\begin{wrapfigure}[18]{r}{0.3\textwidth}
  \begin{center}
    \includegraphics[width=0.3\textwidth]{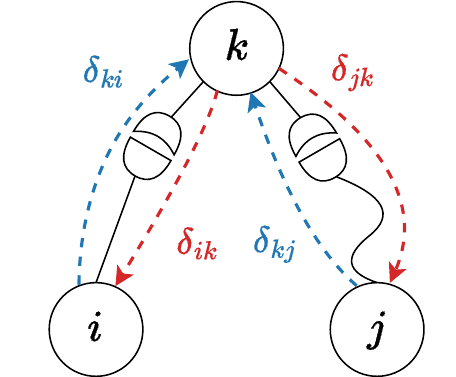}
  \end{center}
  \caption{A three-neuron circuit with delayed forward (blue) and backward (red) signals. Synapses are depicted as a coupling of an axon terminal \img{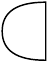} and dendritic spine \img{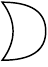}.}
  \label{fig:three-neurons}
\end{wrapfigure}

In CTNNs, communication delays can also result in temporally mismatched signals within the network itself. Consider for example the network in \Cref{fig:three-neurons}, where $\delay{i}{k}$ and $\delay{j}{k}$ represent inter-neuron communication delays of 1 and 2 ms respectively. Imagine that the network is presented with a new input, which is then held fixed. After 1 ms, the signal from neuron $i$ to $k$ will be up-to-date relative to the new input, but the signal that $k$ receives from $j$ will still be carrying information about some previous input. If neuron $k$ updates its activation based on these temporally mismatched signals, its activation will be corrupted, and this corrupted state will influence all downstream computations. Delays can similarly result in corrupted error messages, and therefore disrupt learning as well. 

In this paper, we propose that the disruptive effects of inter-neuron communication delays can be mitigated by having each neuron predict its own future incoming signals based on the signals received up until the present time. This compensation strategy, which we call \textit{prospective messaging} (PM), is local to each neuron in the network and complementary to reducing delay magnitudes directly. Moreover, the choice of prediction model allows us to flexibly balance memory, computation and accuracy requirements. To test this approach, we integrate communication delays and PM in a recently introduced continuous-time deep learning framework called Latent Equilibrium (LE) \citep{haider2021}, which has been used to model cortical microcircuits. We first demonstrate that learning and inference in standard LE networks without PM deteriorate rapidly with the introduction of delays, even if the LE network is massively overparameterized and trained to solve a relatively simple task. We then show that augmenting the LE network with PM recovers performance across a range of delay distributions, network architectures and tasks. 

\section{Communication Delays}
\label{sec:communication_delays}

We begin by formalizing the notion of communication delays, starting with some useful notation. 
Consider an NN with $N$ neurons indexed by the set $[N] := \{1, \dots, N\}$, with input neurons $\mathcal{I} \subset [N]$ and output neurons $\mathcal{O} \subseteq [N] \setminus \mathcal{I}$. The network's parameters consist of a set of weights and biases, with $\Wc{i}{j}{t} \in \mathbb{R}$ denoting the weight of the connection from neuron $i$ to neuron $j$, and $\bi{i}{t} \in \mathbb{R}$ denoting the bias associated with neuron $i$. Each neuron also has an associated activation function $\phi_i$. The environment consists of an $|\mathcal{I}|$-dimensional input vector $\mathbf{x}(t)$ and possibly an instructive signal $\mathbf{y}(t)$. The network's activity is represented by $\usc{1}{t}, \dots, \usc{N}{t}$, where $\usc{i}{t} \in \mathbb{R}$ is the activation of neuron $i$.

In a traditional NN, the activation of a neuron $i \notin \mathcal{I}$ is a function of the signals it receives from its presynaptic neurons. We denote the feed-forward contribution of neuron $i$ to the activation of neuron $j$ as $\ud{i}{j}{t}$. Using this notation, the activation of neuron $j \notin \mathcal{I}$ is typically computed as
\begin{equation}\label{eq:instant-activation}
    \usc{j}{t} = \phi_j\left(\sum_{i=1}^N \Wc{i}{j}{t}\ud{i}{j}{t} + \bi{j}{t} \right).
\end{equation}

The activations of neurons in $\mathcal{I}$ are simply set to the current input $\mathbf{x}(t)$. Under instantaneous inter-neuron communication, each neuron has access to the current (sent) activations of its presynaptic neurons, so $\ud{i}{j}{t} = \usc{i}{t}$. As we have discussed however, instantaneous communication is not possible in physical systems. Therefore, at time $t$, neuron $j$ receives the signal that was sent by each presynaptic neuron $i$ at some past time $t - \delay{i}{j}$, where $\delay{i}{j} \in \R_{> 0}$ denotes the communication delay associated with transmission from neuron $i$ to neuron $j$ (measured in units of time). In this non-instantaneous setting, $\ud{i}{j}{t} = \usc{i}{t - \delay{i}{j}}$. In this work, we assume that delays are static, and discuss the plausibility of this assumption in \Cref{sec:discussion}.

In addition to the forward signals described above, many deep learning algorithms require error signals to be transmitted between neurons. For example, in a network trained via error backpropagation \citep{rumelhart1986}, updates of $\Wc{i}{j}{t}$ directly depend on the adjoints associated with all post-synaptic neurons of neuron $j$ \citep{baydin2018}. If each adjoint is considered to be local to the neuron it is associated with, then it must be transmitted back to neuron $j$ along a feedback connection \citep{whittington2019, lillicrap2020}, and thus incurs a delay.
Following the notation above, we denote sent and received error signals at time $t$ as $\vs{i}{t}$ and $\vs{i}{t-\delay{i}{j}}$, and the value used by neuron $j$ for computation is denoted $\vd{i}{j}{t}$. Collectively, we denote all forward and error signals used by neuron $j$ to update its state by the vector $\zd{j}{t}$. We use $\zs{j}{t - \bm{\delta}_j}$ to denote the signals received by neuron $j$ at time $t$, where $\bm{\delta}_j$ is shorthand for the vector of delays associated with the incoming signals to neuron $j$.

Continuous dynamical systems without delays, such as gradient flow \citep{santambrogio2017, chizat2018, boursier2022} or standard LE networks, which we introduce later, are commonly formalized as ordinary differential equations (ODEs). Introducing delays into such systems turns them into delayed differential equations (DDEs). DDEs are much more complex to characterize and remain less well understood than ODEs~\citep{erneux2009,atay2010}. Even in a one dimensional system with a single feedback loop, infinitely many oscillatory solutions or even chaotic behaviour can be induced by introducing a delay. 

\subsection{Delays Inhibit Learning}
\label{sec:three-neuron-GD}

\begin{figure}[H]
\centering
\begin{subfigure}[b]{0.5\textwidth}
    \includegraphics[width=\textwidth]{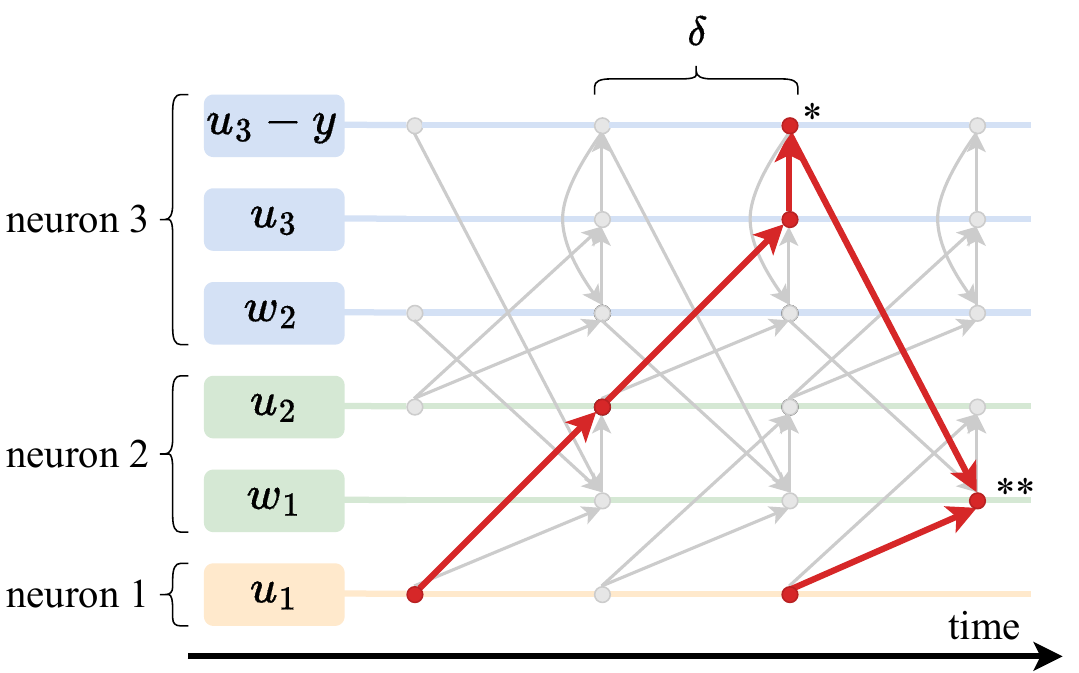}
    \caption{}
    \label{fig:three-neurons-signals}
\end{subfigure}
\hspace{0.5em}
\begin{subfigure}[b]{0.3\textwidth}
    \includegraphics[width=\textwidth]{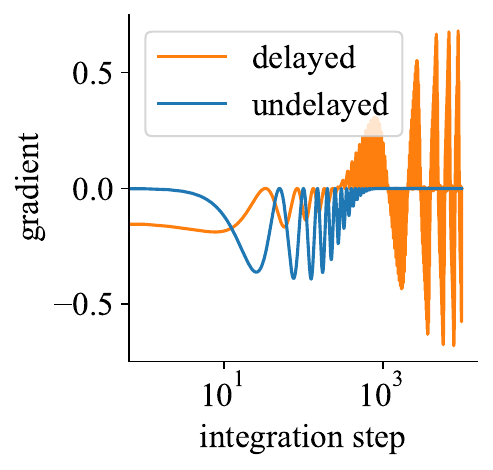}
    \caption{}
    \label{fig:three-neuron-grads}
\end{subfigure}
\caption{\textbf{(a)} Signals propagating through a minimal three-neuron linear network with delays $\delta$. Arrows indicate the transmission of signals between variables being computed continuously through time. Each variable is local to one neuron, indicated by colouring. Red arrows emphasize two signals that converge on $w_1$ in order to compute the gradient with respect to that weight. \textbf{(b)} Gradient dynamics of weight $w_1$. Without delay, gradients converge to the loss minimizers, whereas gradients in the network with delays experience strong oscillations.}
\label{}
\end{figure}

Before introducing our method and the CTNN framework in which we test it, we provide a concrete demonstration of the deleterious effects of inter-neuron communication delays in a traditional streaming gradient descent setting.
Consider a sequence of three neurons connected by scalar weights $ w_1, w_2$, with no biases or activation functions, trained under L2 loss. All signals incur equal delays $\delta$. For simplicity, consider weights to be local to their post-synaptic neurons, and residuals to be computed instantaneously at the output neuron. 
\cref{fig:three-neurons-signals} traces the propagation of signals through the network over time. In red, we highlight the paths of two signals required for the computation of the loss's gradient with respect to $w_1$ at time $t$. We can clearly see the convergence of temporally mismatched information at two crucial points. The first ($*$) occurs as the prediction carrying information about the input at $t-3\delta$ is compared against the environment $y$ at time $t-\delta$. The second mismatch ($**$) incorporates this corrupted residual, which is based on the input at $t-3\delta$, with the input from $t-\delta$. Mismatches such as these occur throughout the network, and disrupt learning. \Cref{fig:three-neuron-grads} shows how the gradient evolves over the course of training. Here, the network  is fed a sine wave with a period of $100$ ms in a streamed fashion. The target signal is identical to the input, and the network's task is to learn the identity function. Without delays, the architecture quickly converges to an optimal minimizer, while the same network under delays of $33$ ms experiences oscillating gradients and does not converge to any value even with an order of magnitude more training. See \cref{app:three-neuron-GD} for additional details.

\section{Prospective Messaging}
\label{sec:prospective_messaging}

In order to avoid lagged or corrupted activations entirely, the signal values used for state updates, i.e. $\zd{j}{t}$, need to equal the sent values $\zs{j}{t}$ for every neuron $j \in N$. As we have discussed, communication delays make this impossible, since the signals received by neuron $j$ at time $t$ carry outdated values $\zs{j}{t - \ddelay{j}}$. To address this issue, we propose a simple delay compensation strategy that maintains neuron-wise locality: have neuron $j$ predict $\zs{j}{t}$ based on the history of signals it has received up until time $t$. We call this approach prospective messaging (PM). Given some signal history $\hist{j}{t} \subseteq \{\zs{j}{t^\prime - \ddelay{j}} : t^\prime \leq t\}$, we need a PM function $\f{j}$ for each neuron in the network, such that $\zd{j}{t} := \f{j}(\hist{j}{t}) \approx \zs{j}{t}$. Crucially, this approach requires that signals in the network carry sufficient conditional mutual information to predict their own future values. We can select any class of predictive model we like for $\f{j}$, providing flexibility in trading off computational and memory complexity for predictive accuracy. We now describe two approaches, one based on linear extrapolation and the other on NNs, which exemplify this trade-off.

\subsection{PM via Linear Extrapolation}
\label{sec:PM-linear_extrapolation}

Given at least approximate knowledge of delays for each connection, one inexpensive option is to predict $\zs{j}{t}$ as a linear extrapolation of the latest signal received by neuron $j$. To illustrate, consider the forward signal from neuron $i$ to neuron $j$, which constitutes one element of $\zd{j}{t}$. Then we have the following prospective message: $\ud{i}{j}{t} = \usc{i}{t - \delay{i}{j}} + \delay{i}{j}\usv{i}{t-\delay{i}{j}}.$
In order to compute this value, we must estimate the instantaneous signal velocity $\usv{i}{t-\delay{i}{j}}$. Using first order finite differences, $\usv{i}{t-\delay{i}{j}} \approx [\usc{i}{t-\delay{i}{j}} - \usc{i}{t - \delay{i}{j} - h}]/h$, where $h$ is some small amount of time. PM via linear extrapolation requires little memory for each neuron, with $\mathbf{h}_j(t) = \{\zs{j}{t - \ddelay{j}}, \zs{j}{t - \ddelay{j} - h}\}$. It is also computationally inexpensive, is local to each synapse, and does not require any learning.

\subsection{PM via Neural Networks}
\label{sec:PM-neural-net}
As we will see in \Cref{sec:experiments}, linear extrapolation has difficulty coping with the complex inter-neuron signals that arise in large networks or volatile environments. Therefore, in order to explore the limits of PM, we investigate PM functions that are themselves NNs. To take advantage of correlations between the forward and error signals converging on a given neuron, we implement a single PM network per neuron. At time $t$, this PM network takes as input $\hist{j}{t} = \{\zs{i}{t - \bm{\delta}_i - \rho_1}, \dots, \zs{i}{t - \bm{\delta}_i - \rho_H}\}$, where $\rho_{1:H}$ are non-negative input lags. Its goal is to predict the true signal $\zs{i}{t}$. The most recent data available at time $t$ to train this network consists of the input $\zs{i}{t - 2\bm{\delta}_i - \rho_1}, \dots, \zs{i}{t - 2\bm{\delta}_i - \rho_H}$ and target $\zs{i}{t - \bm{\delta}_i}$. All of the experiments in \Cref{sec:experiments} use three input lags $\rho_1 = 0, \rho_2 = 10 \cdot dt, \rho_3 = 20 \cdot dt$. In order to train the PM networks in this way, each neuron must again know how much its incoming signals are delayed.

PM functions must perform autoregression on a stream of non-i.i.d. data, received one data point at a time. If the PM function is plastic, it faces an online continual learning task. In particular, PM functions that depend on gradient-based learning, such as NN-based approaches, are vulnerable to catastrophic forgetting \citep{mccloskey1989, parisi2019}. To address this issue, we approximate i.i.d. data by training PM networks on small batches sampled from a memory buffer, rather than only on the most recent available data. Although this approach increases memory costs, it is effective and simple to implement, and therefore serves as an adequate proof of concept. We explore the effects of different memory buffer budgets in \Cref{fig:sawtooth-buffer-sweep}, and suggest more memory-efficient approaches in \Cref{sec:discussion}.

Since the PM network affects the dynamics of the outer LE network, changing it too rapidly can cause instabilities during training. We therefore initialize each PM network with small parameter magnitudes and add a residual connection which adds the network's least-lagged input directly to its output as shown in \Cref{fig:compensation_network}. This ensures that initially the PM network simply outputs its most recent input, and only gradually influences the overall system as it learns to compensate for the delays.
Learning rates for the LE and PM networks are coupled and need to be picked such that the PM network keeps compensating well while allowing the outer LE network to gradually learn the task. Example PM network loss curves are included in \Cref{fig:sawtooth-PM-net-losses}. 

\section{Latent Equilibrium Implementation}
\label{sec:le_implementation}

We investigate the effects of communication delays and test PM in a recently proposed deep learning framework called Latent Equilibrium (LE) \citep{haider2021}. LE networks have continuous-valued signals, and parameters and activations that update (1) in parallel and continuously through time, and (2) using only local information. Moreover, cortical microcircuit models with LE dynamics are the first to execute a valid continuous-time approximation of error backpropagation. These features make LE networks an ideal test bed for PM. 

\begin{figure}[t!]
\centering
\begin{subfigure}[b]{0.48\textwidth}
    \includegraphics[width=\textwidth]{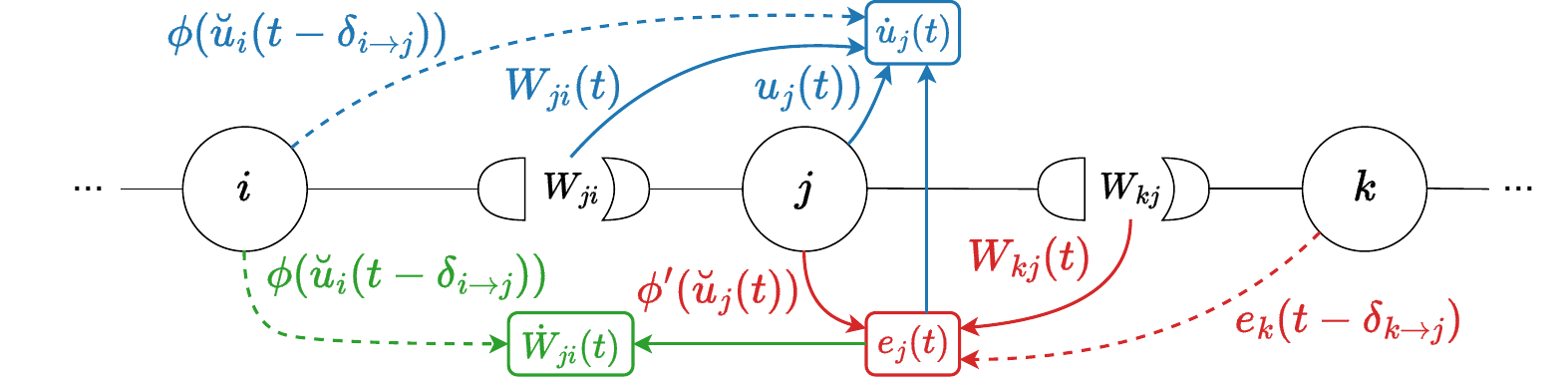}
    \caption{}
    \label{fig:transmitted-signals-hidden}
\end{subfigure}
\hspace{0.5em}
\begin{subfigure}[b]{0.48\textwidth}
    \includegraphics[width=\textwidth]{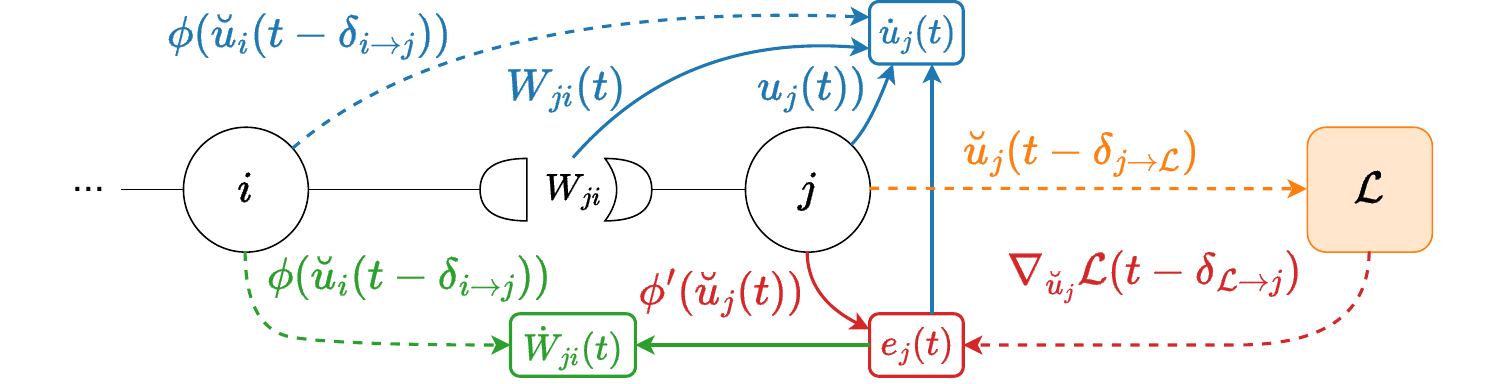}
    \caption{}
    \label{fig:transmitted-signals-output}
\end{subfigure}
\caption[]{Transmission of signals to neuron $j$ as hidden (a) and output (b) neuron in an LE network, for the computation of three states: activation and parameter velocities $\dot{u}_{j}(t)$ and $\Wv{i}{j}{t}$, and error $e_j(t)$. The presynaptic neuron $i$ influences the activation dynamics of neuron $j$ and errors flow back from post-synaptic neuron $k$ or loss node $\mathcal{L}$. Weights are considered local to post-synaptic neurons. Dashed lines represent signals which we consider not to be local to the relevant value being computed, and are therefore delayed. In order to distinguish them from abstract signals (shown as arrows), synaptic connections are depicted as a coupling of an axon terminal \img{figs/axon-terminal.pdf} and dendritic spine \img{figs/dendritic-spine.pdf}.}
\label{fig:transmitted-signals}
\end{figure}

Whereas the present work addresses delays resulting from non-instantaneous inter-neuron signal transmission (i.e. communication delays), LE was developed to address the related problem of non-instantaneous signal integration (i.e. computation delays) resulting from leaky membrane dynamics. To this end, each neuron $j$ in an LE network transmits a linear extrapolation of its current activation downstream. This ``prospective'' activation is defined as $\up{j}{t} := \usc{j}{t} + \tau \usv{j}{t}$, where $\tau \in \R^{+}$ is the membrane time constant.
Since the value sent out by neuron $j$ at time $t$ is now $\up{j}{t}$ instead of $\usc{j}{t}$, all variables $\usc{j}{t}$ and $\usv{j}{t}$ in \Cref{sec:prospective_messaging} are replaced with $\up{j}{t}$ and $\upv{j}{t}$ respectively.

An LE network's dynamics depend on 
error terms which are local to each neuron (see \Cref{fig:transmitted-signals}). The error associated with each hidden neuron $j \notin \{\mathcal{O} \cup \mathcal{I}\}$ is the gradient of a predictive coding-like energy function~\citep{parr2022, millidge2022}, with respect to neuron $j$'s activation. This error value,  
\begin{align}
    \vs{j}{t} = \phi_j^\prime(\up{j}{t})\sum_{k}\Wc{j}{k}{t}\vd{k}{j}{t},
    \label{eq:error-hidden}
\end{align}
depends on the errors of post-synaptic neurons $k$, which must be transmitted backwards. These feedback signals are delayed by $\delay{k}{j}$. For outputs neurons $j \in \mathcal{O}$, the error is the gradient of some loss function $\mathcal{L}$, scaled by $\beta \in \mathbb{R^{+}}$:
\begin{align}
    \vs{j}{t} = - \beta \cdot \vd{\mathcal{L}}{j}{t}.
    \label{eq:error-output}
\end{align}
We assume that the computation of the loss and its gradient are performed by an external module. Therefore, the network's output must be transmitted to this module, and the gradient must be transmitted back, both of which cause delays to be incurred, as depicted in \Cref{fig:transmitted-signals-output}.
Given these errors, the activation and parameter dynamics evolve according to the following two equations: 
\begin{align}
    \tau\usv{j}{t} &= -u_j(t) + \vs{j}{t} + \sum_{i}\Wc{i}{j}{t}\phi_i(\ud{i}{j}{t}) \label{eq:activation-velocity} \\
    \Wv{i}{j}{t} &= \eta_w \cdot \phi_i(\ud{i}{j}{t})\vs{j}{t} \label{eq:weight-velocity}
\end{align}

where $\eta_w \in \mathbb{R}^+$ is the synaptic learning rate. These equations imply that synaptic weight information is instantaneously available to both pre- and post-synaptic neurons. This is a reasonable simplification since synaptic weights are typically considered to change much more slowly than activations \citep{vignoud2022}. However, communication of activations (somatic membrane potentials) to synapses are subject to delays. In particular, since axonal delays are typically larger than dendritic delays \citep{madadiasl2018}, we only implement delays on transmission of pre-synaptic activations to synapses. 
Biases evolve according to $\biv{j}{t} = \eta_b \cdot \vs{j}{t}$, and are considered to be local to their respective neuron. Following \citep{haider2021}, we test a trained model's performance by setting $\beta$ to 0 after a specified number of training iterations, thereby removing the influence of the instructive signal. See \Cref{algo:pseudocode} for pseudocode.

\section{Experiments}
\label{sec:experiments}

In this section, we demonstrate that communication delays disrupt learning in LE networks, and that equipping them with PM recovers performance. 
As we have discussed, PM requires that inter-neuron signals be autocorrelated, so that present signal values can be predicted based on past ones. This in turn means that the input and instructive signals delivered to the LE network must also be autocorrelated. 
We therefore focus our experiments on two classes of tasks where this is the case: Fourier synthesis and streaming video prediction. 
The test losses reported below are computed as the average loss across all time steps after $\beta$ has been set to 0. The integration step size $dt$ is set to $5$ ms. Since we use Euler integration to simulate LE dynamics, we must discretize time, and therefore report delays in units of time steps, i.e. a delay of $\delta$ time steps corresponds to a delay of $\delta \cdot dt$ ms. 
We refer to baseline LE networks without prospective messaging as \textbf{LE}, and LE networks equipped with linear extrapolation-based and NN-based prospective messaging as \textbf{LE-Ex} and \textbf{LE-NN}, respectively. Additional experimental details are provided in \Cref{sec:additional_exp_details}.

\subsection{Fourier Synthesis}
\label{sec:fourier_synthesis}

Fourier synthesis is the process of constructing a complex function from a collection of component sines and cosines (i.e. the inverse of Fourier decomposition). In the following experiments, the LE network receives at time $t$ the input $\mathbf{x}(t) = (\sin{(t;\pi_1)}, \dots, \sin{(t;\pi_{|\mathcal{I}|})})$, where $\pi_{1:|\mathcal{I}|}$ are the periods of the component sine waves, all of which have phase of 0 and amplitude of 1. 

\begin{wrapfigure}[14]{r}{0.4\textwidth}
\centering
\includegraphics[width=0.4\textwidth]{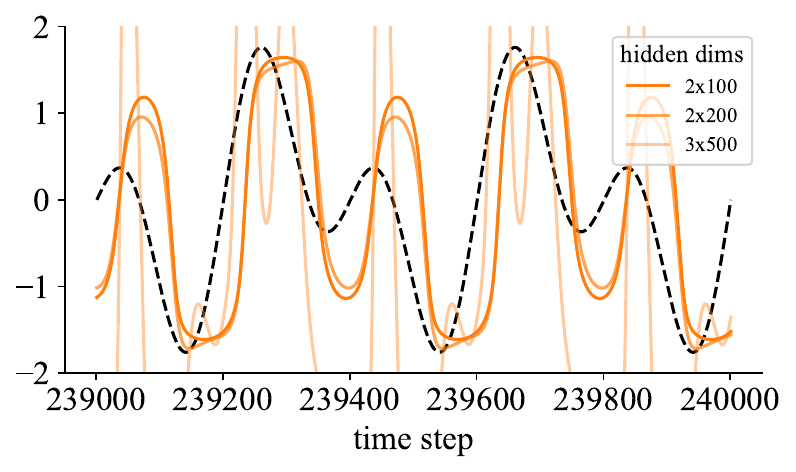}
\caption{Without PM, LE networks with 10-step delays cannot learn the simple two-componet Fourier synthesis task, despite significant overparameterization.}
\label{fig:2-sine-preds-overparam}
\end{wrapfigure}

\begin{figure}[h]
\centering
\begin{subfigure}[b]{0.4\textwidth}
    \includegraphics[width=\textwidth]{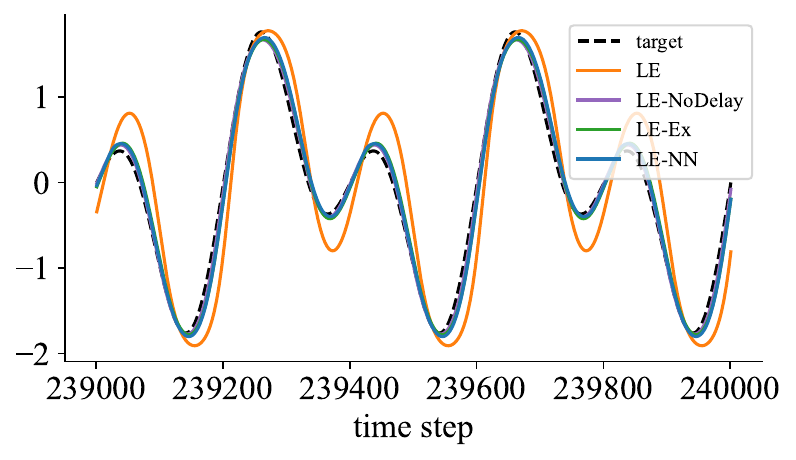}
    \caption{}
    \label{fig:2-sine-preds-stable}
\end{subfigure}
\hspace{0.5em}
\begin{subfigure}[b]{0.4\textwidth}
    \includegraphics[width=\textwidth]{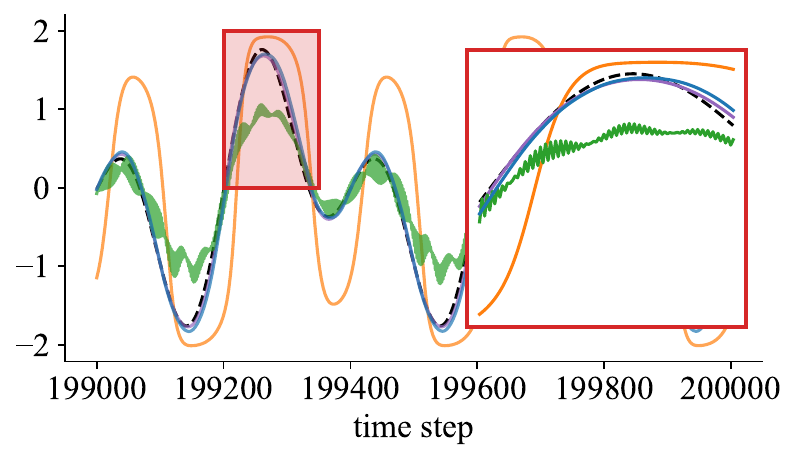}
    \caption{}
    \label{fig:2-sine-preds-unstable}
\end{subfigure}
\caption{Fourier synthesis of two sine waves. Predictions produced by LE without PM (orange), with LE-NN (blue) and with LE-Ex (green), and undelayed LE without PM (purple). \textbf{(a)}  Final 1000 time steps of test period with 5-step communication delays. \textbf{(b)} When delays are increased from 5-steps to 10-steps, linear extrapolation induces high frequency oscillations.}
\label{fig:2-sine-preds}
\end{figure}

We begin with the simple task of adding two sine waves, with periods of 200 and 400 time steps.
We first train an LE networks with one hidden layer of 10 neurons under 5-step delays until convergence, with and without PM. Test-time predictions are depicted in \Cref{fig:2-sine-preds-stable}. We can see that communication delays disrupt the performance of the vanilla LE network, while the addition of PM completely recovers performance to the level of an LE network without delays.

While the efficient LE-Ex from \Cref{sec:PM-linear_extrapolation} works well for this small delay, it requires additional exponential smoothing of $\dot{u}(t)$ in order to avoid oscillations.
Moreover, doubling the delays from 5 to 10 steps without increased smoothing induces strong high frequency oscillations common in DDEs, as seen in \Cref{fig:2-sine-preds-unstable}. We end this section by showing that, even on this simple task, an LE network cannot overcome the effects of delays through overparameterization alone. \Cref{fig:2-sine-preds-overparam} depicts the test-time predictions of LE networks with hidden dimensions (layers $\times$ neurons) of $2\times100$, $2\times200$ and $3\times500$, under 10-step communication delays. In fact, performance deteriorates as the LE network's size is increased. 

\begin{figure}
\centering
\begin{subfigure}[b]{0.4\textwidth}
    \includegraphics[width=\textwidth]{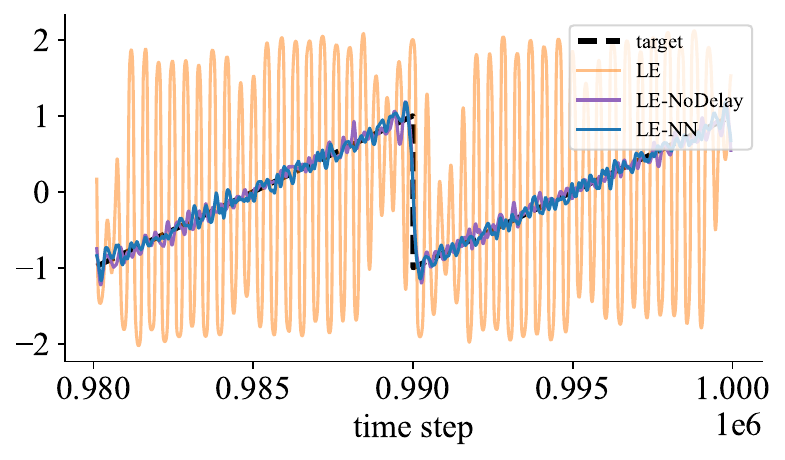}
    \caption{}
    \label{fig:sawtooth-preds}
\end{subfigure}
\hspace{0.5em}
\begin{subfigure}[b]{0.4\textwidth}
    \includegraphics[width=\textwidth]{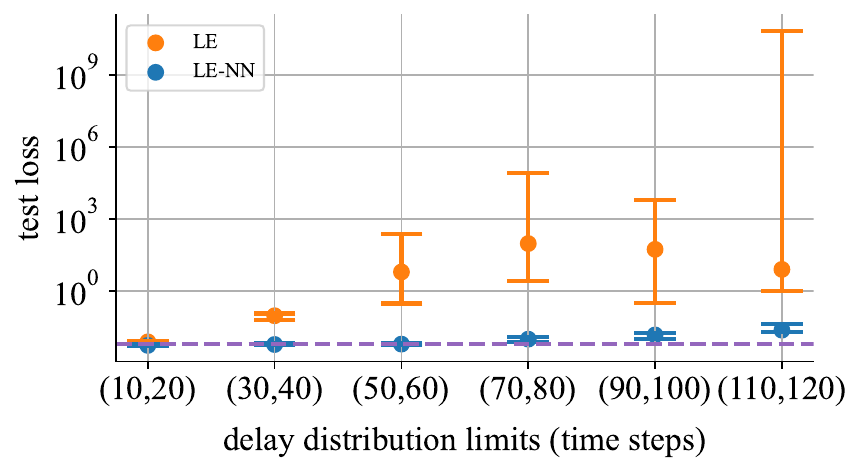}
    \caption{}
    \label{fig:sawtooth-delay-sweep-unif}
\end{subfigure}
\caption{Sawtooth Fourier synthesis for LE networks with one hidden layer of 30 neurons. \textbf{(a)} Test-time predictions with 50 time step communication delays. Basline LE (orange) experiences strong oscillations and does not fit, whereas LE-NN (blue) recovers the performance of an LE network without delays (purple). \textbf{(b)} Sawtooth test loss over a range of uniform delay distributions. Median and range over 5 random seeds shown. The dashed purple line indicates the test loss achieved by an LE network without delays.}
\label{fig:sawtooth}
\end{figure}

Next, we test LE and PM on a more challenging Fourier synthesis task from \citep{haider2021}. Here, the objective is to map 50 input sine components, with evenly spaced periods from 2 ms to 100 ms inclusive, to a sawtooth target with a period of 50 seconds (see \Cref{fig:sawtooth-preds}). This target is especially challenging due its lack of smoothness, and its long period which exacerbates catastrophic forgetting. To cope with the larger input dimension, LE architectures, and delay magnitudes, all sawtooth experiments use PM networks with two hidden layers of 100 neurons.

\begin{wraptable}{r}{7cm}
    \centering
    \caption{Sawtooth test loss over a range of LE network depths (\# hidden layers) and widths (\# neurons per hidden layer), with 50-step communication delays. Mean and standard deviation computed over 5 random seeds.}
    \resizebox{0.5\columnwidth}{!}{
    \begin{tabular}{cc c c c}
        & & \multicolumn{3}{ c }{width} \\
        depth & & 10 & 30 & 50 \\ \cline{1-5}
        \multicolumn{1}{ c }{\multirow{2}{*}{1} } &
        \multicolumn{1}{ c }{LE-NN} & $\mathbf{0.006 \pm 0.001}$ & $\mathbf{0.006 \pm 0.000}$ & $\mathbf{0.006 \pm 0.001}$      \\ \cline{2-5}
        \multicolumn{1}{ c }{} &
        \multicolumn{1}{ c }{LE} & $0.495 \pm 0.271$ & $0.759 \pm 0.258$ & $11.437 \pm 10.992$     \\ \cline{1-5}
        \multicolumn{1}{ c  }{\multirow{2}{*}{2} } &
        \multicolumn{1}{ c }{LE-NN} & $\mathbf{0.021 \pm 0.005}$ & $\mathbf{0.009 \pm 0.001}$ & $\mathbf{0.008 \pm 0.002}$ \\ \cline{2-5}
        \multicolumn{1}{ c }{} &
        \multicolumn{1}{ c }{LE} & $0.133 \pm 0.012$ & $0.124 \pm 0.005$ & $0.133 \pm 0.011$ \\ \cline{1-5}
        \multicolumn{1}{ c }{\multirow{2}{*}{3} } &
        \multicolumn{1}{ c }{LE-NN} & $\mathbf{0.090 \pm 0.012}$ & $\mathbf{0.028 \pm 0.008}$ & $\mathbf{0.023 \pm 0.005}$ \\ \cline{2-5}
        \multicolumn{1}{ c }{} &
        \multicolumn{1}{ c }{LE} & $0.113 \pm 0.009$ & $0.110 \pm 0.006$ & $0.108 \pm 0.008$ \\
    \end{tabular}}
    \label{tab:sawtooth-arch-sweep}
\end{wraptable}
We test performance on a range of delay sizes. Since inter-neuron communication delays in biological networks and neuromorphic chips are heterogeneous, we sample delays from various uniform distributions, while assuming that $\delay{i}{j} = \delay{j}{i}$.
As shown in \Cref{fig:sawtooth-delay-sweep-unif}, adding communication delays to LE networks prevents effective learning, even when delays are small. As delay sizes increase, LE without PM becomes highly unstable, resulting in trumpeting predictions and a divergent loss curves (see \Cref{fig:sawtooth-diverge}). Once NN-based PM is added however, performance is recovered. Sweeps over additional delay distributions are included in \Cref{sec:sawtooth-additional}. We also investigate the effect of 50-step delays on LE networks with different architectures, both with and without NN-based PM. 
The results in \cref{tab:sawtooth-arch-sweep} show that simply increasing LE depth and width does not overcome the negative effects of delays, while PM remains stable and effective as the LE network's size increases.

\subsection{Video Prediction}
\label{sec:video_prediction}

\begin{figure}[t!]
\centering
\begin{subfigure}[b]{0.085\textwidth}
    \includegraphics[width=\textwidth]{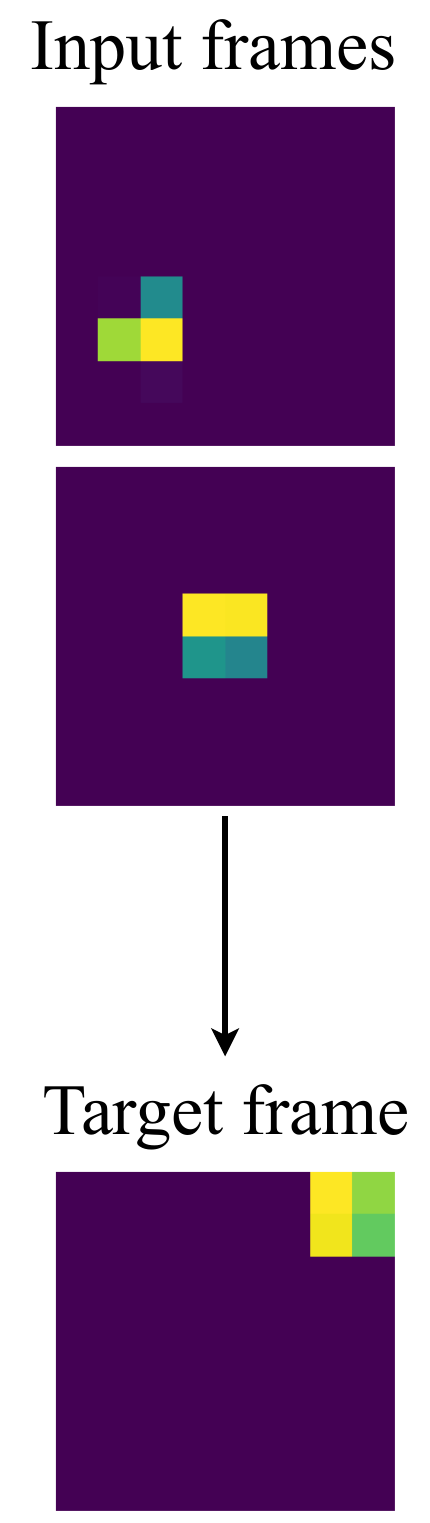}
    \caption{}
    \label{fig:balls-data}
\end{subfigure}
\hspace{0.5em}
\begin{subfigure}[b]{0.6\textwidth}
    \includegraphics[width=\textwidth]{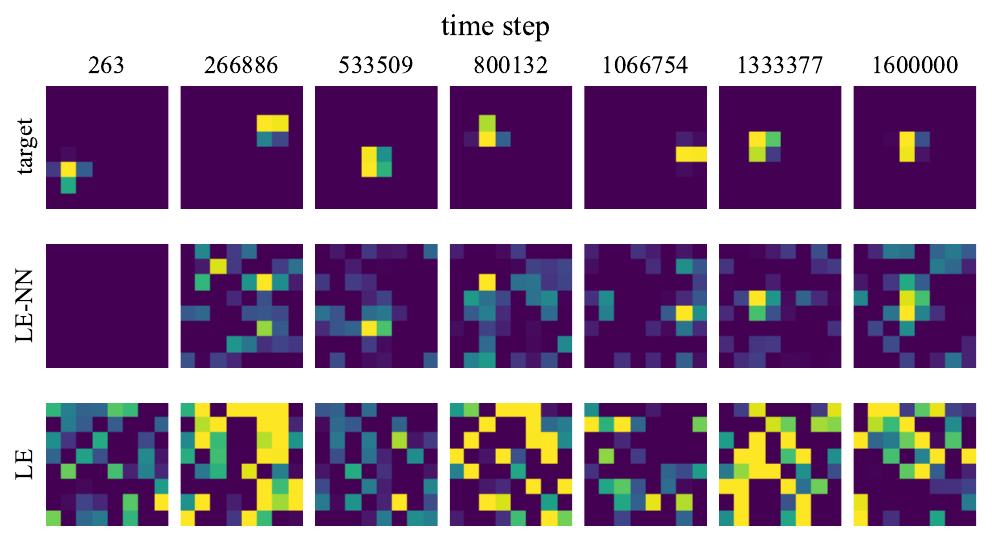}
    \caption{}
    \label{fig:balls-preds}
\end{subfigure}
\caption{\textbf{(a)} Example input-output pair from bouncing ball video frame prediction dataset. Left and right input frames are lagged 800 and 500 time steps from the target frame, respectively. \textbf{(b)} Sample LE-NN and LE predictions over the course of training, along with the corresponding target. Seven approximately equally spaced time steps were chosen from the first step of the training period (after burn-in) to the final step of the test period. $\beta$ set to zero around time step $1.4 \times 10^6$.}
\label{fig:balls}
\end{figure}

In addition to Fourier synthesis, we evaluate our method on an autoregressive video frame prediction task. Similar datasets have been used in the past to evaluate sequence modeling methods \citep{sutskever2008}. Our implementation, adapted from \citep{gan2015}, consists of a long 8x8 video of a ball bouncing deterministically off the sides of the frame without loss of energy. At time $t$, the LE network must predict the current frame given two lagged frames as input. An example input-output pair is depicted in \Cref{fig:balls-data}. This task differs from Fourier synthesis in two important ways: the mapping to be learned is more complex, and the output has a higher dimension. Moreover, this dataset has a longer period ($\sim 45,000$ frames). 

We train an LE network with one hidden layer of 50 neurons on this video prediction task, under 100-step delays. We again use PM networks with two hidden layers of 100 neurons. LE-NN experiences smooth convergence to a relatively low loss, while vanilla LE is highly unstable (see \Cref{fig:balls-loss}). LE-NN achieves a lower loss both at the end of training and during the test phase. In \Cref{fig:balls-preds}, we show a sample of the predictions made by LE and LE-NN over the course of the run. The predictions made by LE-NN are consistently qualitatively superior to LE, which produces higher variance predictions. In particular, the highest-valued pixel in the LE predictions often fails to land on the ball in the target frame. This failure to predict the location of the ball is especially problematic should the model's output be used for object tracking downstream. 

\section{Related Work}
\label{sec:related_work}

Our NN-based PM strategy is similar in spirit to a method for layer-parallel deep learning called synthetic gradients (SG) \citep{jaderberg2017}. Training an NN via error backpropagation typically requires forward and update locking, i.e. enforcement of layer-wise sequential execution of computations during the forward and backward pass prescribed by reverse-mode automatic differentiation. To avoid locking, \citet{jaderberg2017} introduce auxiliary networks which predict forward and gradient signals between layers without having to wait for information to propagate through the network. However, SG differ from PM in several key ways. Most importantly, in order to predict forward signals, SG requires the network's input to be fed directly to each of its layers, whereas PM uses only past signals, which are local to each neuron. Moreover, SG auxiliary networks are shared by all neurons in a layer, again breaking our neuron-local constraint. Finally, we explicitly represent time, and model delays in a unified framework that operates in continuous time without freezing the environment during computation. Another line of work, exemplified by \citet{zheng2017}, addresses stale gradients in model-parallel deep learning by forward-projecting gradients received from workers (e.g. via linear extrapolation) before using them to updating parameters of a global model. Although related to communication delays, these methods focus solely on delayed gradients passed from workers to a global model, rather than delays occurring within the network itself.  

The computational implications of inter-neuron communication delays in biological NN models have been studied for decades.
One extensive area of research is the analysis of these systems from the DDE perspective, characterizing conditions that result in stability, multistability, oscillations, and synchrony in network activity \citep{plant1981, marcus1989, gopalsamy1994, ye1995, vandendriessche1998, shayer2000, buric2003, li2004, wang2006, cao2008, coombes2009, chen2015, xiao2020, tavakoli2021}. Another important area of work is concerned with the role of communication delays in the development and implementation of processes such as sound localization \citep{jeffress1948, carr1988}, motor coordination \citep{pumphrey1938, waxman1971}, and synchronization of signal convergence between various brain regions \citep{pelletier2002, salami2003}.
Moreover, in network models with Hebbian-style plasticity rules, communication delays can induce the emergence of synchronization \citep{gerstner1993, madadiasl2018, aldarabsah2021, madadiasl2022}, polychronization \citep{izhikevich2006}, temporal feature maps \citep{kempter2001} and neural ensembles \citep{kerr2013}. The manipulation of delays via local cooling has also been used as a tool to investigate the function of various brain circuits \citep{xu2013, banerjee2021}. Finally, modifiable axonal delays have been proposed as an alternative learning mechanism to traditional synaptic plasticity \citep{seidl2010, seidl2014, hussein2014, wang2015, fields2015, matsubara2017, cullen2021, sun2022}.

In light of the computational benefits and drawbacks of communication delays, it has been suggested that delay compensation mechanisms co-evolved with mechanisms that utilise delays to implement computations \citep{nijhawan2009}. One direct approach to mitigating the effects of communication delays is to decrease their magnitudes, especially by increasing axonal conduction speeds. Animals have evolved numerous adaptations to this end, axon myelination being the most notable \citep{hartline2007}. However, increasing conduction speed is resource-intensive. In monkey brains, the fastest axons achieve $\sim400\times$ the conduction speed of the slowest axons, but at the cost of occupying $\sim40,000\times$ the volume per unit length \citep{swadlow2012}. In fact, myelin alone accounts for approximately 50$\%$ of the mass of a human brain \citep{salzer2016}. Therefore, a number of computational compensation strategies have also been proposed.

Among these prior works, the most similar to ours is that of \citet{lim2006b}. These authors propose to compensate for delays through ``facilitating neural dynamics,'' similar to the linear extrapolation method in \Cref{sec:PM-linear_extrapolation}. However, they use an evolutionary algorithm to learn their network's parameters, thereby avoiding the stability issues that can arise as the input-output mapping changes during gradient-based learning. Moreover, they employ a fully recurrent network architecture wherein all input features are fed directly to every hidden and output neuron. Only these input connections incur delays, whereas in our work delays exist for all internal connections. Finally, their method only works under relatively small delays, and requires an initial period without delays to ensure stability. A primary motivation of \citep{lim2006b} is to provide a computational account of the flash-lag effect (FLE), a family of psychological phenomena wherein moving objects are perceived ahead of their true positions \citep{mackay1958, nijhawan2002}. Analogous anticipatory effects have also been observed in motor behaviour \citep{nijhawan2003}. Several other works aiming to explain FLE through delay compensation mechanisms are similar to PM in spirit \citep{nijhawan2009, khoei2017}, but are generally restricted to network topologies and neuron dynamics that are specific to FLE, and do account for the learning of these features. 

\section{Discussion}
\label{sec:discussion}

In this paper, we present a general computational approach to overcoming the effects of inter-neuron communication delays in continuous-time NNs (CTNNs). After formalizing a general notion of communication delays, we introduce a delayed version of Latent Equilibrium (LE), a state-of-the-art CTNN framework \citep{haider2021}. We then propose a delay compensation approach, called prospective messaging (PM), wherein each neuron tries to compensate for delays by predicting the signals it would be receiving if communication was instantaneous, given the delayed signals it has received so far. We implement two PM strategies, which exemplify either end of the  resource-performance trade-off. The first, linear extrapolation, requires minimal memory and computation, but can be unstable. On the other hand, NN-based PM is successful in difficult settings, at the cost of significant memory and computation requirements. In Fourier synthesis and next-frame prediction tasks, we show that delays disrupt learning in LE networks, and that their effects are alleviated by the integration of PM. We validate our results on a range of delay distributions and LE network architectures. 

Our method represents one way in which physically implemented NNs can mitigate the harmful effects of delays. PM is biologically plausible insofar as it requires only neuron-local information, and operates in continuous time. On the other hand, NN-based PM requires considerable computation to be performed within each neuron. Recent work has suggested that single neurons can in fact execute complex operations \citep{beniaguev2021, stelzer2021}. Since these computations are performed via intra-cellular molecular mechanisms, they may not incur delays as large as those experienced during inter-cellular communication. Nonetheless, more efficient PM strategies should be developed. Moreover, communication delays in biological networks are not static. Action potential initiation sites are not constant \citep{melinek1996}, and axonal conduction velocities are history dependent \citep{decol2012, zhang2017, chereau2017}. This is not reflected in this paper. The primary obstacle to the practical use of NN-based PM is the necessity of a replay buffer to combat catastrophic forgetting. This can be improved by replacing replay with more memory efficient continual learning strategies, such as those which leverage context-dependent parameter regularization \citep{kirkpatrick2017}. However, an efficient solution to catastrophic forgetting remains a major open problem in machine learning, so we leave alternative continual learning strategies to future work.

\bibliography{references.bib}
\bibliographystyle{unsrtnat_initials.bst}

\newpage
\appendix

\section{Experimental Details}
\label{sec:additional_exp_details}

\subsection{Hyperparameters}

During the training phase, $\beta$ is set to $0.1$ in all experiments. Following \citet{haider2021}, activations $\phi_j$ are set to $\tanh$ for all hidden neurons, and identity for all input and output neurons. 

\begin{table*}[h]
    \centering
    \caption{Hyperparameters for experiments presented as figures in the paper. Note that smoothing factors of 1.0 correspond to no smoothing.}
    \resizebox{\columnwidth}{!}{
    \begin{tabular}{llllllll}
        \toprule
        LE parameter & \cref{fig:2-sine-preds-stable} & \cref{fig:2-sine-preds-unstable} & \cref{fig:2-sine-preds-overparam} & \cref{fig:sawtooth-preds} & \cref{fig:sawtooth-delay-sweep-unif} & \cref{tab:sawtooth-arch-sweep} & \cref{fig:balls-preds}\\
        \midrule
        \makecell[l]{Hidden layers dims \\ (layers $\times$ neurons)} & $1 \times 10$ & $1 \times 10$ & -- & $1 \times 30$ & $1 \times 30$ & -- & $1 \times 50$\\
        Learning rate & $0.1$ & $0.1$ & $0.05$ & $0.05$ & $0.05$ & $0.05$ & $0.05$ \\
        Delay (time steps) & $5$ & $10$ & $10$ & $50$ & -- & $50$ & $100$ \\
        Smooth factor \; $\bar{\mathbf{s}}$ & $1.0$ & $1.0$ & $1.0$ & $1.0$ & $1.0$ & $1.0$ & $1.0$ \\
        Smooth factor \; $\dot{u}$ & $1.0$ & $1.0$ & $1.0$ & $1.0$ & $1.0$ & $1.0$ & $1.0$\\
        \toprule
        PM-Ex parameter & & & \\
        \midrule
        Finite diff. step size ($h$) & $1 \cdot dt$ & $1 \cdot dt$ & -- & -- & -- & -- & -- \\
        Smooth factor \; $\bar{\mathbf{s}}$ & $1.0$ & $1.0$ & -- & -- & -- & -- & -- \\
        Smooth factor \; $\dot{u}$ & $0.5$ & $0.5$ & -- & -- & -- & -- & -- \\
        \toprule  
        PM-NN parameter & & & \\
        \midrule
        Batch size & $1$ & $1$ & -- & $5$ & $5$ & $5$ & $10$\\
        Buffer size (time steps) & $500$ & $500$ & -- & $10^4$ & $10^4$ & $10^4$ & $4 \times 10^4$\\
        Learning rate & $0.002$ & $0.002$ & -- & $0.002$ & $0.002$ & $0.002$ & $0.002$\\
        Initial parameter gain & $0.1$ & $0.1$ & -- & $0.1$ & $0.1$ & $0.1$ & $0.1$\\
        Smooth factor \; $\bar{\mathbf{s}}$ & $0.5$ & $0.5$ & -- & $0.5$ & $0.5$ & $0.5$ & $0.5$\\
        Smooth factor factor \; $\dot{u}$ & $1.0$ & $1.0$ & -- & $1.0$ & $1.0$ & $1.0$ & $1.0$\\
    \end{tabular}}
\end{table*}

\subsection{Resources}

In this section, we document the compute resources used for the experiments presented in \cref{sec:experiments}. All runs for \cref{fig:2-sine-preds-stable,fig:2-sine-preds-unstable} were performed locally an 8-Core CPU (Intel Core i9) in under 15 minutes. All runs for other experiments in \cref{sec:experiments} were performed on a 20-core CPUs (Intel Xeon E5-2650v3), with 30-CPU multi-threading. Runs for 2x100, 2x200 and 3x500 LE networks described in \cref{fig:2-sine-preds-overparam} took 10 minutes, 15 minutes and 3 hours respectively. All runs for \cref{fig:sawtooth} took under 10 minutes for LE baselines, and 2.5 hours for LE-NN. The bouncing balls experiments in \cref{fig:balls} took approximately 45 minutes to run for the LE baseline, and 55 hours for LE-NN. Finally, run times for LE and LE-NN experiments for the architecture sweep presented in \cref{tab:sawtooth-arch-sweep} are recorded in \cref{tab:sawtooth-arch-sweep-resources}.

\begin{table}[h]
    \centering
    \caption{Compute time (hours:minutes) for experiments in \cref{tab:sawtooth-arch-sweep}, rounded to nearest minute.}
    \label{tab:sawtooth-arch-sweep-resources}
    \begin{tabular}{cc c|c|c}
        & & \multicolumn{3}{ c }{width} \\
        depth & & 10 & 30 & 50 \\ \cline{1-5}
        \multicolumn{1}{ c }{\multirow{2}{*}{1} } &
        \multicolumn{1}{ c }{LE-NN} & 1:18 & 2:18 & 3:09      \\ \cline{2-5}
        \multicolumn{1}{ c }{} &
        \multicolumn{1}{ c }{LE} & 0:06 & 0:10 & 0:12     \\ \cline{1-5}
        \multicolumn{1}{ c  }{\multirow{2}{*}{2} } &
        \multicolumn{1}{ c }{LE-NN} & 1:57 & 4:15 & 7:16 \\ \cline{2-5}
        \multicolumn{1}{ c }{} &
        \multicolumn{1}{ c }{LE} & 0:08 & 0:16 & 0:22 \\ \cline{1-5}
        \multicolumn{1}{ c }{\multirow{2}{*}{3} } &
        \multicolumn{1}{ c }{LE-NN} & 2:40 & 6:19 & 11:22 \\ \cline{2-5}
        \multicolumn{1}{ c }{} &
        \multicolumn{1}{ c }{LE} & 0:12 & 0:23  & 0:31  \\
    \end{tabular}
\end{table} 

\clearpage
\section{Method Details}

\subsection{Streaming Gradient Descent with Delays}
\label{app:three-neuron-GD}
In \cref{sec:three-neuron-GD}, we illustrate how temporally mismatched signals can arise in minimal networks under delayed streaming gradient descent dynamics. The network is comprised of three serially-connected neurons, connected by two scalar weights $w_1$ and $w_2$, which are considered to be local to their post-synaptic neurons. All forward and backward signals in the network incur equal delays of $\delta$.
The activation $u_1(t)$ of the input neuron is set to $x(t)$, while the hidden and output neurons update their values according to delayed forward messages: 
\begin{align*}
    u_2(t) &= w_1(t) \cdot u_1(t-\delta),\\
    u_3(t) &= w_2(t) \cdot u_2(t-\delta).
\end{align*}
The network is trained under an $L2$ loss, so weight gradients depend on a residual $r(t) = u_3(t) - y(t)$. We assume this value to be computed locally at neuron $3$, which has instantaneous access to the target signal. The gradient of the loss with respect to the parameters is computed as follows:
\begin{align*}
    \nabla_{w_2}\mathcal{L}(t) &= r(t) \cdot u_2(t-\delta),\\
    \nabla_{w_1}\mathcal{L}(t) &= r(t-\delta) \cdot w_2(t-\delta) \cdot u_1(t-\delta).
\end{align*}

\subsection{PM Network Architecture}
\label{app:pm-network-architecture}

\begin{figure}[H]
  \begin{center}
    \includegraphics[width=0.3\textwidth]{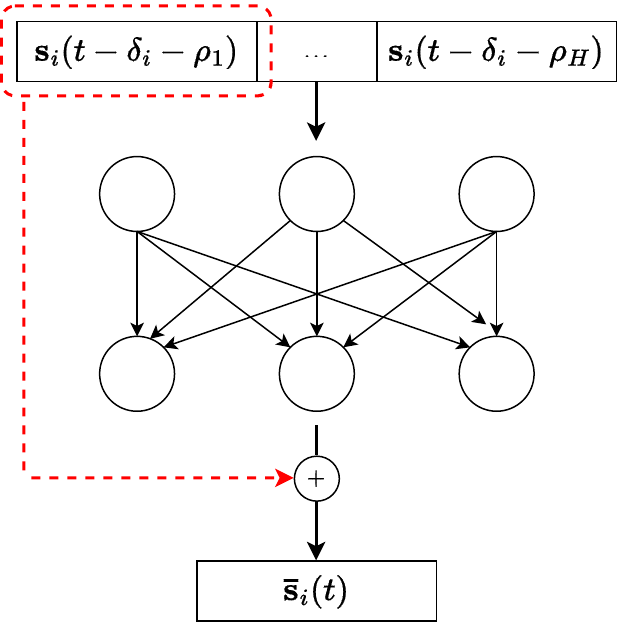}
  \end{center}
  \caption{A schematic of a neural network-based prospective messaging function. The most recent input to the network is added to the network's output via a residual connection, as shown by the red dashed line. In conjunction with low initial parameter magnitudes, this promotes stable dynamics in the outer LE network.}
  \label{fig:compensation_network}
\end{figure}

\clearpage


\subsection{Pseudocode}
\label{app:pseudocode}
\begin{algorithm}[ht]
\caption{Pseudocode for LE-PM integration}\label{alg:high_level}
\begin{algorithmic}[1]
\STATE $t \leftarrow 0$
\WHILE{$t < t_{\text{max}}$}
    \IF{$t > t_{\beta_{\text{off}}}$}
        \STATE $\beta \leftarrow 0$ 
    \ENDIF
    \STATE Update input neuron activations and loss module with $\mathbf{x}(t)$ and $\mathbf{y}(t)$    
    \FOR{each neuron $n \in N \setminus \mathcal{I}$}
        \STATE Update neuron $n$'s signal history trace with $\zs{n}{t - \bm{\delta}_n}$
    \ENDFOR
    \STATE Update loss module's signal history trace with $\zs{\mathcal{L}}{t - \bm{\delta}_\mathcal{L}}$
    \FOR{each non-input layer in the neural network}
        \FOR{each neuron $n$ in the layer}
            \STATE Apply PM: \; $\zd{n}{t} \leftarrow f_n(\hist{n}{t})$
            \IF{PM plasticity is active}
                \STATE Update PM mechanism $f_n$
            \ENDIF
            \STATE Compute error signal (Eq. \ref{eq:error-hidden} or \ref{eq:error-output})
            \STATE Compute activation and parameter velocities (Eq. \ref{eq:activation-velocity} and \ref{eq:weight-velocity})
            \STATE Compute next membrane potential: $u_n(t + dt) \leftarrow u_n(t) + dt \cdot \dot{u}_n(t)$
            \STATE Compute next prospective potential: $\breve{u}_n(t + dt) \leftarrow u_n(t) + \tau \cdot \dot{u}_n(t)$
        \ENDFOR
        \IF{last layer}
            \STATE Apply PM: \; $\zd{\mathcal{L}}{t} \leftarrow f_\mathcal{L}(\hist{\mathcal{L}}{t})$
            \STATE Update PM mechanism $f_\mathcal{L}$
        \ENDIF
        \IF{LE plasticity is active}
            \STATE Update weights and biases for the layer
        \ENDIF
    \ENDFOR
    \STATE $t \leftarrow t + dt$
\ENDWHILE
\end{algorithmic}
\label{algo:pseudocode}
\end{algorithm}

\clearpage
\section{Additional Experimental Results}

\subsection{Fourier Synthesis -- Sawtooth}
\label{sec:sawtooth-additional}
\begin{figure}[H]
\centering
\begin{subfigure}[b]{0.48\textwidth}
    \includegraphics[width=\textwidth]{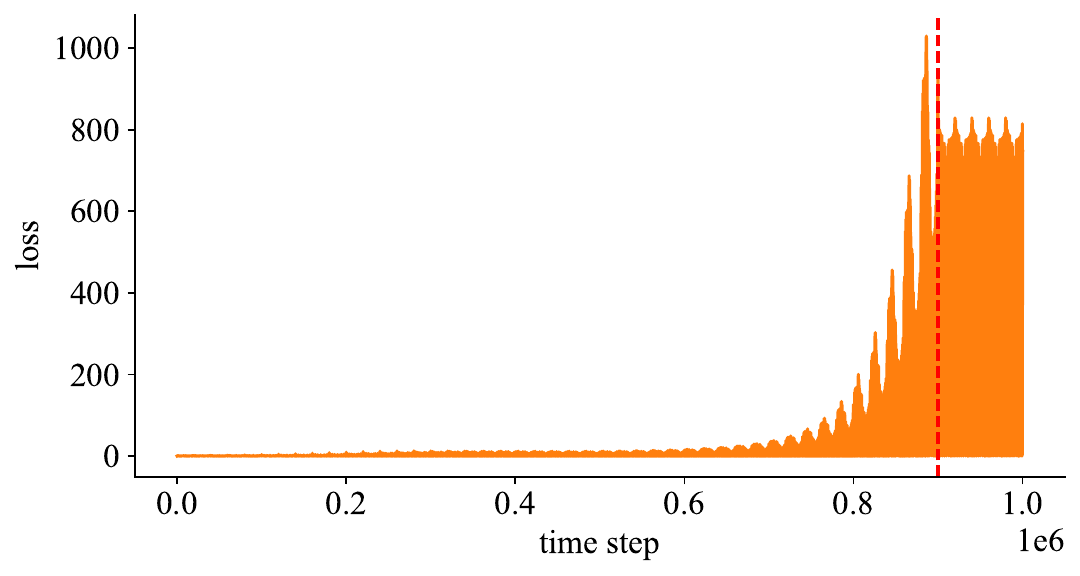}
    \caption{}
    \label{fig:sawtooth-diverge-loss}
\end{subfigure}
\hspace{0.5em}
\begin{subfigure}[b]{0.48\textwidth}
    \includegraphics[width=\textwidth]{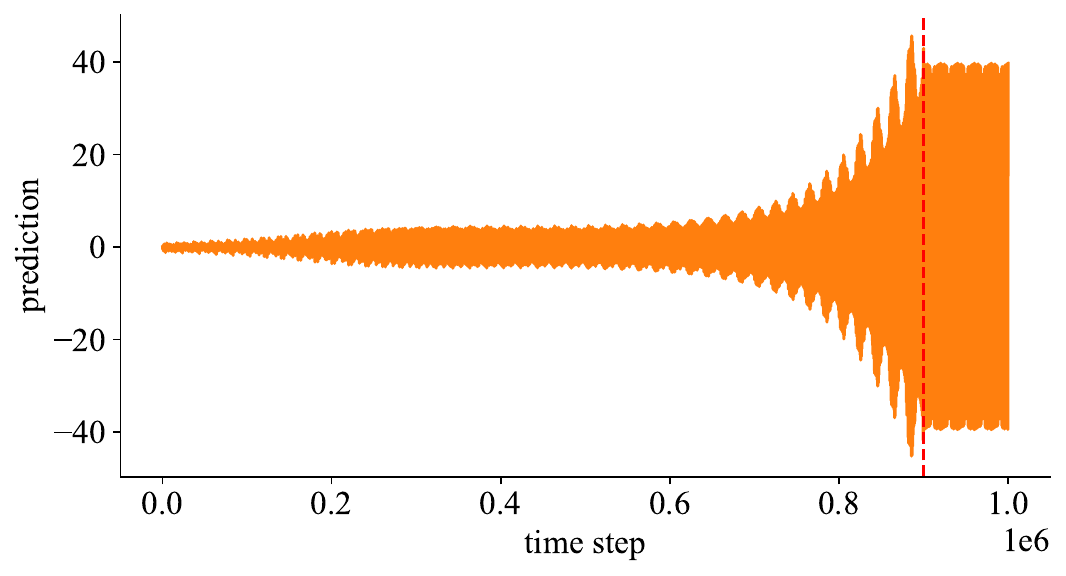}
    \caption{}
    \label{fig:sawtooth-diverge-preds}
\end{subfigure}
\caption{Diverging loss \textbf{(a)} and trumpeting predictions \textbf{(b)} resulting from training an LE network without PM on the sawtooth-target Fourier synthesis task, under homogeneous delays of 75 time steps. This LE network had one hidden layer of 30 neurons. The dashed red lines indicate the time step where $\beta$ is set to zero.}
\label{fig:sawtooth-diverge}
\end{figure}

\begin{figure}[H]
\centering
\begin{subfigure}[b]{0.48\textwidth}
    \includegraphics[width=\textwidth]{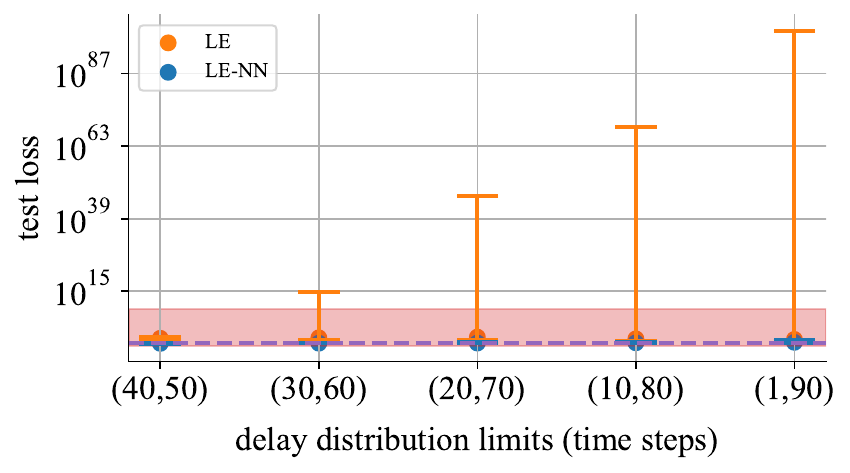}
    \caption{}
\end{subfigure}
\hspace{0.5em}
\begin{subfigure}[b]{0.48\textwidth}
    \includegraphics[width=\textwidth]{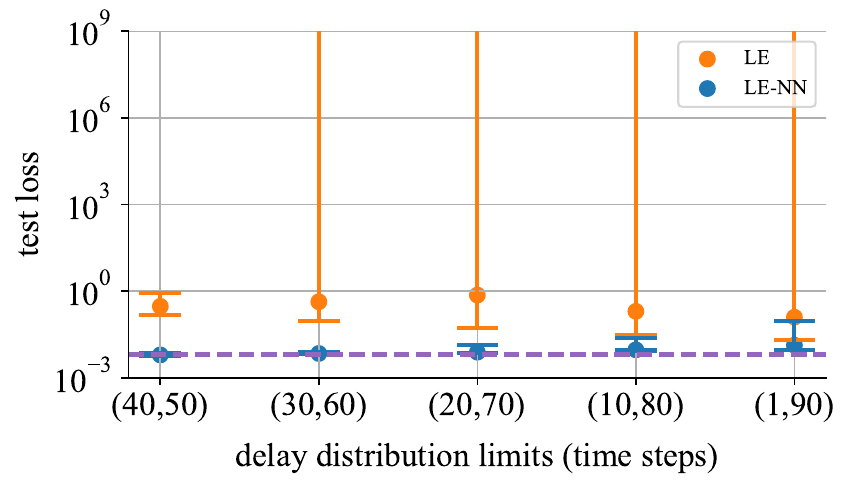}
    \caption{}
\end{subfigure}
\caption{Sawtooth test loss over a range of uniform delay distributions. LE network has one hidden layer of 30 neurons. In this case, the sweep is over the variance of the delay distribution. The red area in \textbf{(a)} indicates the zoomed in region depicted in \textbf{(b)}. Median and range over five random seeds shown. The dashed purple line indicates the test loss achieved by an LE network without delays or PM.}
\label{fig:sawtooth-delay-sweep-unif-var}
\end{figure}

\begin{figure}[H]
\centering
\includegraphics[width=0.7\textwidth]{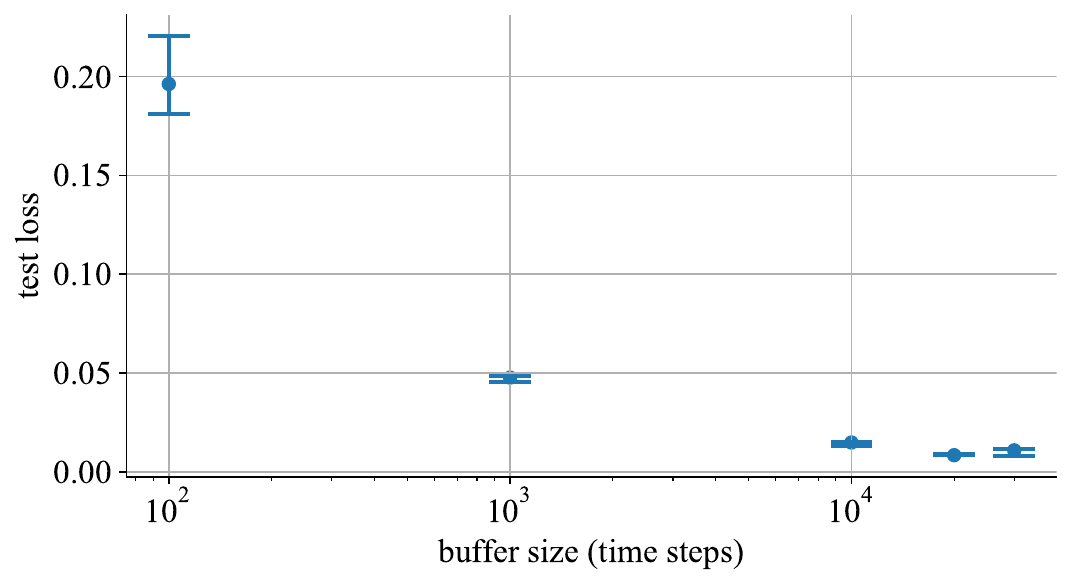}
\caption{Median and range of test performances over a range of memory buffer sizes, for five random seeds. There is a clear trend in favor of buffer sizes on the order of $10^4$ time steps, which is the period of the sawtooth target. We also tested buffer sizes of 10 and 0 (i.e. only using the most recently available training pair), but these resulted in massive numerical instabilities.}
\label{fig:sawtooth-buffer-sweep}
\end{figure}

\begin{figure}[ht]
    \centering
    \includegraphics[width=0.9\columnwidth]{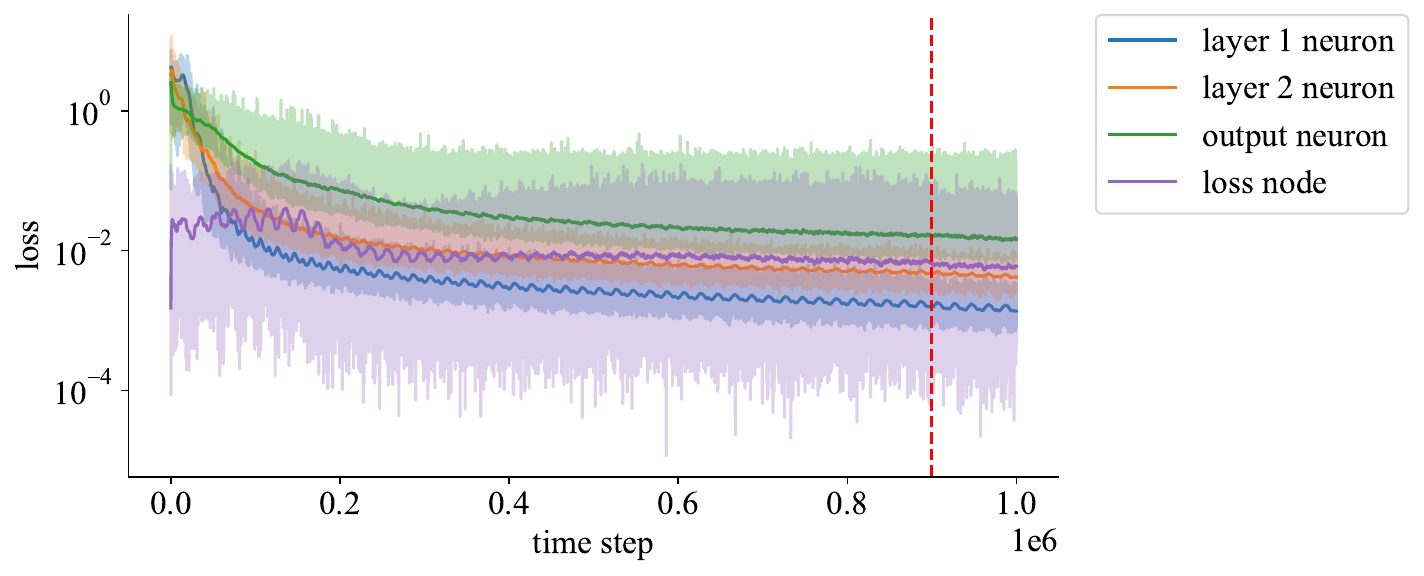}
    \caption{Loss curves for the PM network of one neuron from each layer an LE network with two hidden layers of 50 neurons, as well as the loss node. Communication delays are 50 time steps. Darker lines depict the smoothed version of the corresponding lighter ones for clarity. Dashed red line indicates the beginning of the test phase. While the loss curves of the first two layers converge smoothly in the first third of training to a high precision and stably predict the incoming messages, the output neuron does not achieve the same precision and has higher variance. The loss node achieves high precision as well, but oscillates in the beginning and increases after the initial descent before it decreases again. Such behavior is often observed during training the compensation networks, in particular in the last layers. As the outer LE network is learning the PM-NN needs to catch up and can temporarily fall behind in training as described in~\Cref{sec:PM-neural-net}.}
    \label{fig:sawtooth-PM-net-losses}
\end{figure}

\subsection{Video Prediction}

\begin{figure}[H]
\centering
\includegraphics[width=0.7\textwidth]{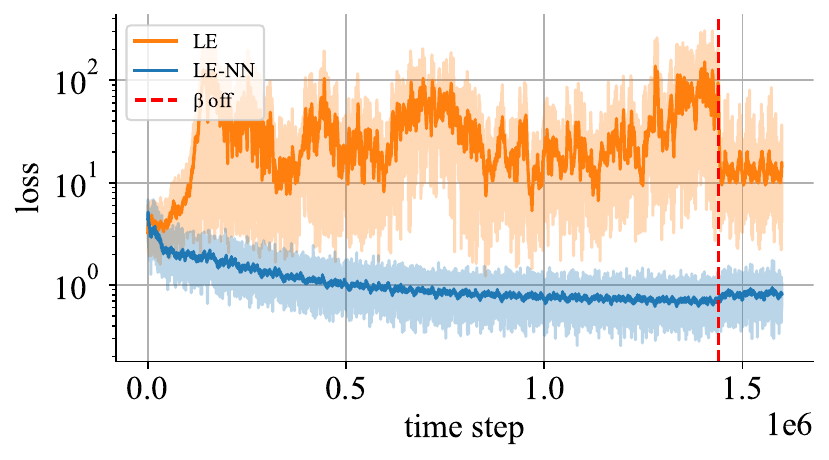}
\caption{Loss curves from the bouncing balls experiment presented in \Cref{sec:video_prediction}. The darker lines depict the smoothed version of the corresponding lighter ones for clarity. The dashed red line indicates the time step where $\beta$ is set to zero, i.e. the start of the test period.}
\label{fig:balls-loss}
\end{figure}

\section{Broader Impacts}
\label{app:broader-impacts}


Coping with communication delays can help to scale (deep) learning systems and increase their reach into new domains, e.g. through implementation in neuromorphic hardware. Resulting systems could amplify both the benefits, such as more efficiency in industrial systems and augmentation of the human intellect, as well as the dangers of (deep) learning systems, some of which are AI safety, discriminatory biases and disruption of industries and job markets. While our work could have an effect on any of these aspects, it can be considered foundational research for now and no immediate societal impacts are obvious to us.

\end{document}